\documentclass[letterpaper]{article} 
\usepackage{aaai25}  
\usepackage{times}  
\usepackage{helvet}  
\usepackage{courier}  
\usepackage[hyphens]{url}  
\usepackage{graphicx} 
\usepackage{natbib}  
\usepackage{caption} 
\usepackage[dvipsnames]{xcolor}

\newcommand{\etc}{\emph{etc. }}
\newcommand{\ie}{\emph{i.e. }}
\usepackage{multirow}
\usepackage{dblfloatfix}
\usepackage{subcaption}
\usepackage{amsmath}
\usepackage{amsfonts}
\usepackage[ruled]{algorithm2e}
\usepackage{tcolorbox}
\usepackage{booktabs}
\usepackage{cuted}

\usepackage{xcolor}
\definecolor{orange}{HTML}{f4b083}
\definecolor{verylightgray}{gray}{0.9} 

\usepackage{listings}
\title{Interpretable Failure Detection with Human-Level Concepts}
\author{
    Kien X. Nguyen, Tang Li, Xi Peng
}
\affiliations{
    Department of Computer and Information Sciences\\
    University of Delaware\\

    Newark, DE, USA\\
    \{kxnguyen, tangli, xipeng\}@udel.edu
}

\usepackage{bibentry}

\begin{document}

\maketitle


\begin{figure*}
    \centering
    \includegraphics[width=0.6\linewidth]{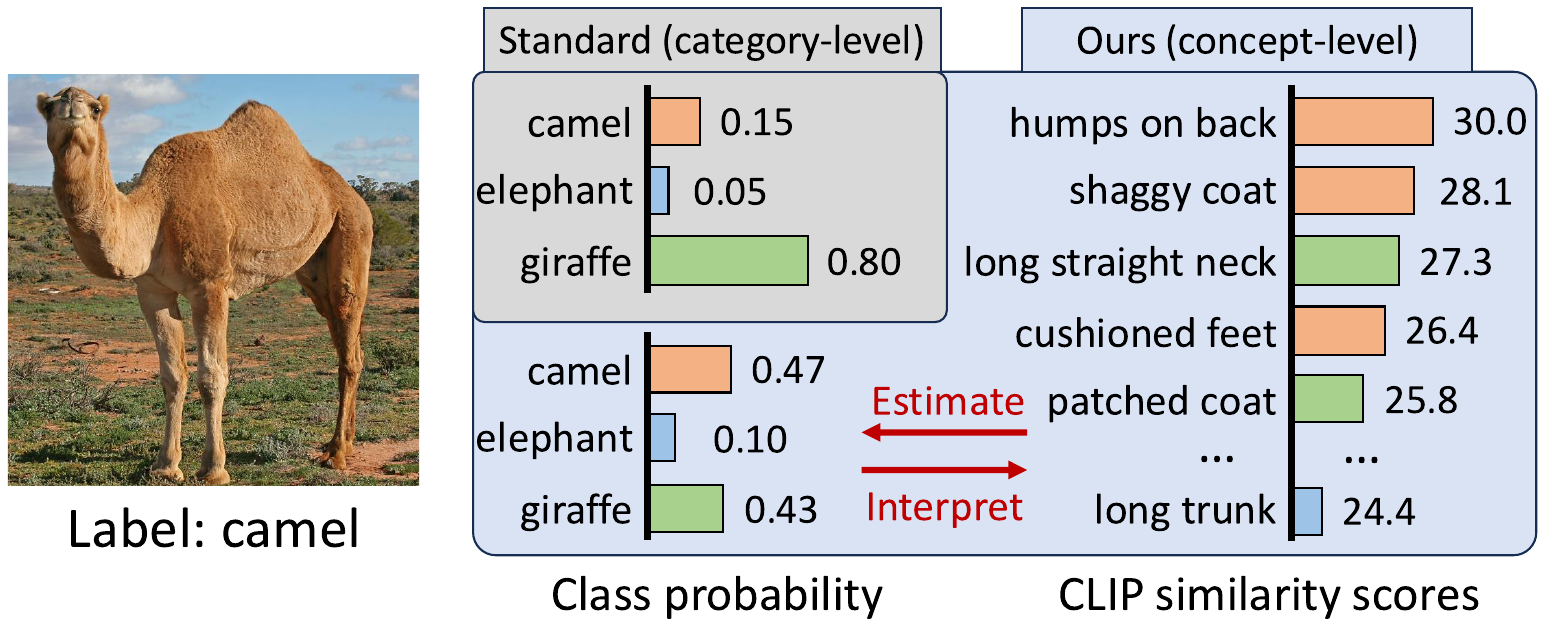}
    \captionof{figure}
    {
    Comparison between standard (MSP) and our approaches. MSP relies solely on class logits to predict failures, which is problematic in detecting overconfident but incorrect predictions. To tackle this problem, we propose to deconstruct each category into its associated human-level concepts for a \textit{finer-grained} estimate of confidence. 
    }
    \label{fig:title-figure}
\end{figure*}

\begin{abstract}
Reliable failure detection holds paramount importance in safety-critical applications.
Yet, neural networks are known to produce overconfident predictions for misclassified samples. 
As a result, it remains a problematic matter as existing confidence score functions rely on category-level signals, the logits, to detect failures. 
This research introduces an innovative strategy, leveraging human-level concepts for a dual purpose: to reliably detect \textit{when} a model fails and to transparently interpret \textit{why}.
By integrating a nuanced array of signals for each category, our method enables a finer-grained assessment of the model's confidence.
We present a simple yet highly effective approach based on the ordinal ranking of concept activation to the input image. 
Without bells and whistles, our method significantly reduce the false positive rate across diverse real-world image classification benchmarks, specifically by $3.7\%$ on \textit{ImageNet} and $9\%$ on \textit{EuroSAT}.
\end{abstract}
    
\section{Introduction}
Vision-language models have demonstrated impressive capability across diverse visual recognition domains~\cite{Radford2021LearningTV,Jia2021ScalingUV,Singh2021FLAVAAF,Li2022BLIPBL, Li2023BLIP2BL}.
However, when it comes to safe deployment in high-stake applications, it is of paramount importance for a model to be self-aware of its own shortcomings.
For instance, in monitoring for natural disasters such as floods or wildfires, the AI system must signal for human intervention upon encountering scenarios where its confidence is low. 
Such self-awareness ensures that preemptive measures can be taken to mitigate disaster impacts on communities and ecosystems.
Therefore, it is imperative not only to detect failures accurately but also to understand the reasons behind them.

Traditional methods~\cite{Hendrycks2016ABF, Granese2021DOCTORAS, Zhu2023OpenMixEO, Zhu2023RethinkingCC, liang2020enhancing} rely on category-level information to detect misclassifications, performing confidence estimation on the class logits.
However, neural networks are known to produce overconfident predictions for misclassified samples due to factors like spurious correlations~\cite{Arjovsky2019InvariantRM, Sagawa2019DistributionallyRN}, thus existing confidence scoring functions (CSFs) fall short in such cases.
Besides, the model confidence depicted through category-level information impedes the ability for humans to interpret \textit{why} it fails. 
To this end, we ask the following question: \textit{``What other sources of information can we leverage to enhance failure detection?"}


We present a novel perspective on detecting failures by leveraging human-level concepts, or visual attributes.
With the flexibility to incorporate free-form language to VLMs (\ie CLIP), we can represent a category with a set of predefined concepts~\cite{Menon2022VisualCV, Oikarinen2023LabelFreeCB,li2024beyond,li2024deal}.
Instead of only prompting the model \textit{``Do you recognize a camel?"}, we collectively ask \textit{``Do you recognize humps on back?''}, or \textit{``Do you recognize shaggy coat?"}.
The purpose is to measure the model's confidence in the object's detailed visual attributes in addition to the holistic category.
We thus achieve a more \textit{accurate} confidence estimate to detect failures more effectively (Fig.~\ref{fig:title-figure})

Ideally, a VLM that can recognize a image of a camel should also recognize all the associated visual attributes, such as \textit{humps on back}, \textit{shaggy coat}, \etc
Such visual attributes should yield higher confidence scores compared to those associated with the absent categories.
Conversely, if the model shows high confidence in concepts from multiple unrelated categories at the same time, it could indicate a failure in its recognition process.
Based on such intuition, we present a simple but effective approach using the \textbf{O}rdinal \textbf{R}anking of \textbf{C}oncept \textbf{A}ctivation (ORCA) to detect failures.
Additionally, these human-understandable concepts allow users to understand the reasons behind such failures, thereby aiding them in refining the training process.

We rigorously validate our method's efficacy in detecting incorrect samples across both natural and remote sensing image benchmarks, which mirror the complexity in real-world scenarios. 
ORCA demonstrates a significant capability to mitigate the issue of overly confident misclassifications.
In summary, our contributions are threefold:
\begin{enumerate}
    \item We leverage human-level concepts to detect \textit{when} and interpret \textit{why} a model fails using vision-language models.
    \item We present a simple but effective approach, called ORCA, to estimate more reliable confidence via the ordinal ranking of the concepts' activation.
    \item We empirically demonstrate that the concept-based methods enhance failure prediction performance across a wide range of classification benchmarks.
\end{enumerate}
\section{Related Work}
\textbf{Failure Detection.}
Failure detection, or misclassification detection, is a burgeoning area of research within the realm of artificial intelligence. 
Detecting when machine learning models produce incorrect or misleading predictions has significant implications for safety, reliability, and transparency in various domains.
Existing research in this field falls into two main categories: (1) retraining or fine-tuning of neural networks~\cite{Moon2020ConfidenceAwareLF,Zhu2023OpenMixEO,Zhu2023RethinkingCC}, and (2) the design of novel confidence score functions~\cite{Granese2021DOCTORAS,Hendrycks2016ABF}. 
The former approach involves retraining or fine-tuning neural networks with specific objectives aimed at improving the model's capability to recognize its own failures.
Zhu et al.~\cite{Zhu2023RethinkingCC} employs a training objective that seeks flat minima to mitigate overconfident predictions. While these approaches have shown promise, they often require extensive computational resources and access to the entire model, which may not be feasible for large VLMs.
Researchers have also turned their attention to the design of new CSFs~\cite{Granese2021DOCTORAS}.
Despite these efforts, the most robust CSF remains the MSP~\cite{jaeger2023a}. 
However, a downside of category-level CSFs is their inability to detect overconfident but incorrect predictions, which is problematic. 
In this work, we deconstruct category-level into concept-level signals to achieve a more nuanced estimate of the model's confidence.

A closely related sub-field is \textit{confidence calibration}~\cite{minderer2021revisiting,levine2023enabling,Mukhoti_2020_NIPS,pereyra2017regularizing}, where the goal is to adjust a model's predicted probabilities to ensure that they accurately reflect the true likelihood of those predictions being correct.
However, Zhu et al.~\cite{Zhu2023RethinkingCC} has empirically shown that calibration methods frequently yield no benefits or even detrimentally affect failure prediction.
Similarly, some works~\cite{jaeger2023a,Bernhardt2022FailureDI} also emphasizes the importance of \textit{confidence ranking} over \textit{confidence calibration} in failure detection. 
Some other related sub-fields are \textit{predictive uncertainty estimation}~\cite{pmlr-v48-gal16, BlundellJMLR15, LakshminarayananNIPS17, Mukhoti_2023_CVPR}, \textit{out-of-distribution detection}~\cite{Zhu2023OpenMixEO, liang2020enhancing, pmlr-v180-dinari22a,lee2018simple}, \textit{open-set recognition}~\cite{vaze2022openset, Geng_2021} and \textit{selective classification}~\cite{geifman2017selective,fisch2022calibrated}.

\vspace{5pt}
\noindent\textbf{Human-level Concepts in Vision-Language Models.}
Concept-based models (CBMs) aim to open the black box of neural networks. 
Concept bottleneck networks are pioneers for interpretable neural networks, with each neuron in the concept bottleneck layer representing a concept at the human level~\cite{Koh2020ConceptBM, Yuksekgonul2022PosthocCB}.
With the flexibility to employ free language in vision language models, such as CLIP~\cite{Radford2021LearningTV}, ALIGN~\cite{Jia2021ScalingUV}, FLAVA~\cite{Singh2021FLAVAAF}, and BLIP~\cite{Li2022BLIPBL, Li2023BLIP2BL}, concepts of human level can be naturally integrated into the prediction mechanism~\cite{Menon2022VisualCV, Yang2022LanguageIA, Oikarinen2023LabelFreeCB}.
This work can be viewed as a variant of CBMs for failure detection, which has never been considered before.
We show the approach better predicts failures and, as a byproduct, helps interpret \textit{why} a model fails.
\section{Backgrounds}

\textbf{Overview on Failure Detection. } We consider failure detection on the multi-class classification task. 
Let $\mathcal{X} \in \mathbb{R}^d$ be the input space and $\mathcal{Y} = \{ 1, 2, \dots, C \}$ be the label space, where $d$ is the dimension of the input vector.
Given a data set $\{ (\mathbf{x}_i, y_i) \}_{i=1}^N$ with $N$ data points independently sampled from the joint probability distribution $\mathcal{X} \times \mathcal{Y}$, a standard neural network $f: \mathcal{X} \rightarrow \mathcal{Y}$ outputs a probability distribution over the $C$ categories.
For an input $\mathbf{x}$, $f$ outputs $\boldsymbol{\hat{p}} = \hat{P}(y|\mathbf{x};\theta)$ as the class probabilities, where $\theta$ denotes the network's parameters.
In the context of failure detection, we consider a pair of functions $(f, g)$, where $g: \mathcal{F} \times \mathcal{X} \rightarrow \mathbb{R}$ is the confidence scoring function, and $f \in \mathcal{F}$.
With a predefined threshold $\tau \in \mathbb{R}^+$, the failure detection output is defined as:
\begin{equation}
    (f,g)(\mathbf{x}) = 
    \begin{cases}
   \hat{P}(y|\mathbf{x};\theta), & \text{if } g(f, \mathbf{x}) \geq \tau \\
    \text{detect},              & \text{otherwise.}
\end{cases}
\end{equation}

Failure detection is initiated when $g(f, \mathbf{x})$ falls below a threshold $\tau$. 
Ideally, a confidence scoring function should output higher confidence scores for correct predictions and lower confidence scores for incorrect predictions.
Despite efforts in designing CSFs, Jaeger et al.~\cite{jaeger2023a} has shown that the standard Maximum Softmax Prediction remains the best CSF across a wide range of datasets and network architectures. 
Mathematically, MSP is defined as:
\begin{equation}
    g(f, \mathbf{x}) = \max_{c\in\mathcal{Y}} \hat{P}(y=c|\mathbf{x};\theta)
\end{equation}
which returns the maximum output signal after the softmax activation function on the network output layer.

\noindent\textbf{Failure Detection with VLM.} \label{subsec:zs}
CLIP~\cite{Radford2021LearningTV}, a vision-language model, is pre-trained on a large-scale dataset comprising of 400 million image-text pairs. 
CLIP uses contrastive learning to align the image and text pairs. 
During inference, we calculate the model's logits as the cosine similarity score between the input image embedding and the corresponding text embeddings. 
Given an input image $\mathbf{x}$, the embedding is denoted as $f_{\text{img}}(\mathbf{x}) \in \mathbb{R}^m$. 
In addition, $C$ text labels represent the category names $\{ \mathbf{t}_c \}_{c=1}^C$, where $f_{\text{txt}}(\mathbf{t}_c) \in \mathbb{R}^m$ are the embeddings and $m \ll d$. 
For each category, we calculate the corresponding logit as:
\begin{equation} \label{eq:simscore}
    s_c = 100 \times \frac{f_{\text{img}}(\mathbf{x}) \cdot f_{\text{txt}}(\mathbf{t}_c)}{\lVert f_{\text{img}}(\mathbf{x}) \rVert \lVert f_{\text{txt}}(\mathbf{t}_c) \rVert}
\end{equation}
where $\lVert \cdot \rVert$ is the $L_2$ norm. 
The softmax function then converts the logits into probabilities:
\begin{equation} \label{eq:softmax}
    \hat{p}_c = \frac{\exp({s_c})}{\sum_{j=1}^{C} \exp({s_j})}
\end{equation}
where $\hat{p}_c \in \boldsymbol{\hat{p}}$. $f(\mathbf{x}) = \operatorname{argmax}_{c \in \mathcal{Y}} \hat{p}_c$ is the prediction, and $g(f, \mathbf{x}) = \max_{c \in \mathcal{Y}} \hat{p}_c$ can be regarded as the model confidence for a given input $\mathbf{x}$ using MSP. 
\section{Methods}
Traditional methods rely on the category-level signals to estimate the model's confidence.
This leads to unreliable confidence estimate as neural networks are prone to overconfident misclassification. 
To address this issue, we suggest exposing the model to diverse viewpoints via human-level concepts. 
Rather than inquiring about the model's certainty regarding an image being a camel, we also query its confidence regarding specific attributes like the presence of humps on the camel's back, a shaggy coat, \etc

\begin{figure*}[!t]
    \centering
    \includegraphics[width=\textwidth]{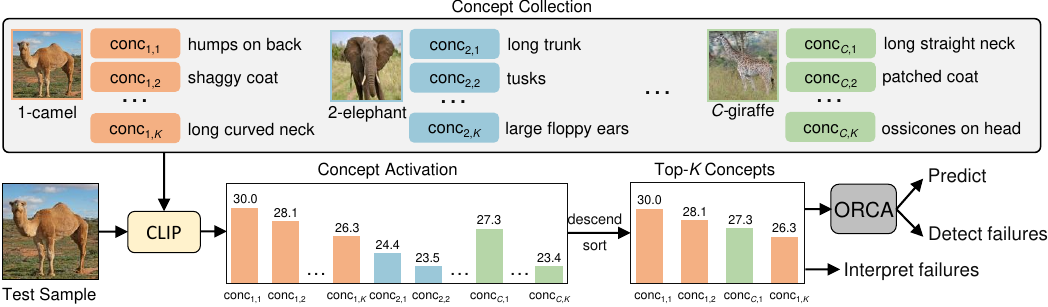}
    \caption{Overview of the ORCA framework. We first prompt GPT-3.5 to construct the concept collection $\mathcal{A}$. We then pass the image and all the concepts into CLIP to retrieve the concept similarity scores, represented by the number above each bar, and sort them in descending order. Based on the top-$K$ responses, we analyze the interaction among concept activations through ordinal ranking to predict the model's failures, and interpret why it fails. ``Detect failures'' is triggered when the confidence falls below a predefined threshold. Best viewed in color.}
    \label{fig:overview}
\end{figure*}

Recent advancements in VLMs enable such integration of human-level concepts as free-form language into the pipeline~\cite{Menon2022VisualCV, Yang2022LanguageIA, Oikarinen2023LabelFreeCB}. 
In this section, we describe the integration of the work by Menon and Vondrick~\cite{Menon2022VisualCV} which employs concept aggregation to establish a baseline concept-based method for failure detection. Subsequently, we introduce ORCA, our novel approach that captures the interaction among concept activations through ordinal ranking, enhancing the reliability of failure detection.
\subsection{Human-Level Concepts for Failure Detection} \label{sec:desc-clip}
Given $K$ concepts per category, we define $\mathcal{A}$ as a collection of all concepts, where $|\mathcal{A}| = C \times K$. 
We obtain the vector of similarity scores (or logits), $S_{\text{conc}} = [s_{1,1}, \dots, s_{1,K}, s_{2,1}, \dots, s_{C, K}]$, between the image embedding and all the concepts using Eq.~\ref{eq:simscore}.
DescCLIP then calculates the mean similarity score among all concepts for each category $c$ to retrieve the logits and output the prediction:
\begin{equation}
    f(\mathbf{x}) = \operatorname{argmax}_{c\in\mathcal{Y}} \frac{1}{K} \sum_{k=1}^K s_{c, k}
\end{equation}

Finally, we apply the softmax function (Eq.~\ref{eq:softmax}) on the logits to get the class probabilities and employ MSP to obtain the model's confidence score.

\subsection{Ordinal Ranking of Concept Activation}

DescCLIP's concept aggregation leads to a coarse-grained confidence estimation procedure. We propose a fine-grained approach that models the interaction among concepts via ordinal ranking to estimate confidence more reliably. 

Ideally, if a model is confident about predicting a category $\hat{c}$ then the concepts associated with $\hat{c}$ should yield the strongest activations. 
In other words, the similarity scores of all concepts belonging to $\hat{c}$, $\{s_{\hat{c}, k}\}_{k=1}^K$, should belong to the top-$K$ ranking.
Conversely, we would see a mixture of concepts from different categories in the top-$K$ ranking if the model is likely to make an incorrect prediction. 
With such information, we can separate correct and incorrect predictions more reliably.
Next, we describe two variants of our proposed method: baseline and rank-aware ORCA.
In brevity, the former builds upon simple counting mechanisms, while the latter weighs the concept contributions to the confidence estimate based on their ranks.

\vspace{5pt}
\noindent\textbf{Baseline ORCA.}
We first sort $S_{\text{conc}}$ in descending order and retrieve the set of the top-$K$ concepts, denoted as an ordered set $\mathcal{A}_{\text{top-}K}$.
After that, we derive the confidence based on the number of different categories whose concepts belong in $\mathcal{A}_{\text{top-}K}$.
The rationale is straightforward: the model is at a higher risk of failure as there are more categories featuring in $\mathcal{A}_{\text{top-}K}$.
The prediction is determined as follows:
\begin{equation} \label{eq:orca-b-pred}
    \begin{aligned}
    f(\mathbf{x}) &= \operatorname{argmax}_{c\in\mathcal{Y}} \lvert \mathcal{A}_{\text{top-}K} \cap \mathcal{A}_c \rvert,
\end{aligned}
\end{equation}
where $\mathcal{A}_c$ denotes the set of concepts of an arbitrary category $c$'s concepts, and $\lvert \cdot \rvert$ denotes the set cardinality.
The confidence of the prediction is the ratio between the number of the predicted category's concepts appearing in $\mathcal{A}_{\text{top-}K}$ over $K$:
\begin{equation} \label{eq:orca-b-conf}
    g(f, \mathbf{x}) = \frac{\lvert \mathcal{A}_{\text{top-}K} \cap \mathcal{A}_{\hat{c}}\rvert}{K} 
\end{equation}
where $\hat{c} = f(\mathbf{x})$ is the prediction. 
We dub this variant ORCA-B in the text.



\noindent\textbf{Rank-aware ORCA.}
While ORCA-B provides a fundamental approach, its reliance solely on rudimentary counting mechanisms limits its ability to capture nuanced distinctions.
To enhance our approach, we introduce a rank-aware variant that uses ordinal ranking information to deliver more accurate failure detection.
In detail, we construct a rank-aware weight vector $\mathbf{w}$ where the value of each element is proportional to the ordinal ranking.
First, we define the ordinal ranking vector $\mathbf{r} = [K, K-1, \dots, 1]$ with $K$ elements in descending order.
Then, we apply a logarithmic weighting function to assign each rank in $\mathbf{r}$ a weight $w_i \in \mathbf{w}$, resulting in a decreasing vector whose elements sum up to $1$. 
Logarithmic ensures a smooth distribution of weights among the ranks of each concept, enabling a more nuanced estimation of the confidence level.
Specifically, the logarithmic scaling equation is defined as $w_i = \frac{\log(1 + r_i)}{\sum_{j=1}^{K} \log(1 + r_j)}$, with the normalization of each weight $w_i$ in $\mathbf{w}$. 
Finally, for each category $c$ with its concepts featuring in $\mathcal{A}_{\text{top-}K}$, we calculate the prediction and the confidence of the model as follows:
\begin{equation} \label{eq:orca-r-pred}
    f(\mathbf{x}) = \operatorname{argmax}_{c\in\mathcal{Y}} \sum_{k=1}^K \mathbb{I}(a_k \in \mathcal{A}_c) \cdot w_k 
\end{equation}
\begin{equation} \label{eq:orca-r-conf}
    g(f, \mathbf{x}) = \operatorname{max}_{c\in\mathcal{Y}} \sum_{k=1}^K \mathbb{I}(a_k \in \mathcal{A}_c) \cdot w_k 
\end{equation}

\noindent where $a_k$ is the $k^{\text{th}}$ concept in the ordered set $\mathcal{A}_{\text{top-}K}$, and $\mathbb{I}(\cdot)$ denotes the indicator function that returns $1$ if the condition is true. We refer to this variant as ORCA-R.
\begin{table*}[!htb]
    \centering
    \resizebox{0.8\linewidth}{!}{
    
    \begin{tabular}{llcccccc}
    \toprule
          \multirow{2}{*}{\textbf{Dataset}}&\multirow{2}{*}{\textbf{Method}}& \multicolumn{3}{c}{ResNet-101} & \multicolumn{3}{c}{ViT-B/32}\\
          \cmidrule(l{3pt}r{3pt}){3-5} \cmidrule(l{3pt}r{3pt}){6-8}
          &&\textbf{AUROC} $\uparrow$ & \textbf{FPR95} $\downarrow$ & \textbf{ACC} $\uparrow$ & \textbf{AUROC} $\uparrow$ & \textbf{FPR95} $\downarrow$ &\textbf{ACC} $\uparrow$  \\
    
    \cmidrule(l{3pt}r{3pt}){1-5} \cmidrule(l{3pt}r{3pt}){6-8}
          \multirow{7}{2cm}{CIFAR10 ($K = 10$)}&Zero-shot $+$ MSP & 85.98 & 62.98 & 78.01 & 88.92&  58.66 & 88.92\\
          &\hspace{1.35cm} $+$ ODIN & 83.65 & 65.50 & 78.01 & 84.49&   65.36& 88.92\\
          &\hspace{1.35cm} $+$ DOCTOR & 86.56 & 63.76 & 78.01 & 88.58&   62.32& 88.92\\
          & Ensemble $+$ MSP & \textbf{86.35} & 63.53 & 80.97 & \textbf{89.25} & 57.03 & 89.70\\
          & \hspace{1.4cm} $+$ ODIN &83.39 & 67.95& 80.97 &  83.66& 63.34& 89.70\\
          & \hspace{1.4cm} $+$ DOCTOR & 85.67 & 66.53 & 80.97 &  88.68& 58.87& 89.70\\
          & DescCLIP $+$ MSP & 85.84 & 64.68 & 80.70 &  89.28 &  58.77 & 88.80\\
          & \hspace{1.45cm} $+$ ODIN & 80.92 & 68.34 & 80.70 & 82.61 & 66.83 & 88.80 \\
         & \hspace{1.45cm} $+$ DOCTOR & 84.99 & 67.92 & 80.70 & 88.80 & 61.64 & 88.80 \\
          \cmidrule(l{3pt}r{3pt}){2-5} \cmidrule(l{3pt}r{3pt}){6-8}
          &ORCA-B & 84.90 & 66.09 & \underline{80.98} &  87.34 &  \textbf{50.52}& \underline{89.34}\\
          &ORCA-R & \underline{85.93} & \textbf{62.68} & 80.60 & \underline{89.00} &  \underline{52.70} & \textbf{90.00}\\
    
    \cmidrule(l{3pt}r{3pt}){1-5} \cmidrule(l{3pt}r{3pt}){6-8}
         \multirow{7}{2cm}{CIFAR100 ($K = 20$)}& Zero-shot $+$ MSP & 80.72 & 73.40 & 48.50 & 81.15 & 71.09 & 58.42\\
         & \hspace{1.35cm} $+$ ODIN & 77.21& 75.13& 48.50 & 76.93& 71.08& 58.42\\
         & \hspace{1.35cm} $+$ DOCTOR & 79.68& 75.36& 48.50 & 81.57& 69.40& 58.42\\
         & Ensemble $+$ MSP & 79.22 & 73.43 & 48.66 & 81.44 & 70.88 & 63.91 \\
         & \hspace{1.4cm} $+$ ODIN & 75.59& 76.00& 48.66 & 75.73& 73.87& 63.91 \\
         & \hspace{1.4cm} $+$ DOCTOR & 77.96& 76.47& 48.66 & 80.02&74.06& 63.91 \\
         & DescCLIP $+$ MSP & 80.22 & 73.39 & 52.90 & 82.54 &  67.38 & 66.70 \\
         & \hspace{1.45cm} $+$ ODIN & 75.86 & 75.35 & 52.90 & 75.72 & 73.11 & 66.70 \\
         & \hspace{1.45cm} $+$ DOCTOR & 79.09 & 74.96 & 52.90 & 81.30 & 70.83 & 66.70 \\
         \cmidrule(l{3pt}r{3pt}){2-5} \cmidrule(l{3pt}r{3pt}){6-8}
         & ORCA-B & 80.35 & \textbf{70.46} & 52.16 & \underline{83.35} & \underline{67.35} & 66.00\\
         & ORCA-R & \textbf{80.46} & \underline{72.38} & \textbf{53.11} & \textbf{83.40} & \textbf{67.00} & \underline{66.50}\\

    \cmidrule(l{3pt}r{3pt}){1-5} \cmidrule(l{3pt}r{3pt}){6-8}
         \multirow{7}{2cm}{ImageNet ($K = 25$)}& Zero-shot $+$ MSP & 78.93 & 74.05 & 56.67 & 79.44 & 72.91 & 58.37 \\
         & \hspace{1.35cm} $+$ ODIN & 70.59& 80.75& 56.67 & 70.48& 80.07& 58.37 \\
         & \hspace{1.35cm} $+$ DOCTOR & 78.38&  75.90& 56.67 & 79.01& 74.17& 58.37 \\
         & Ensemble $+$ MSP & 78.58 & 74.37 & 56.73 & 79.66 & 72.89 & 59.22 \\
         & \hspace{1.4cm} $+$ ODIN & 70.29&  80.98& 56.73 &  70.61& 80.55 & 59.22 \\
         & \hspace{1.4cm} $+$ DOCTOR & 77.98 & 76.25 & 56.73 & 78.34 & 76.24 & 59.22 \\
         & DescCLIP $+$ MSP & 80.09 & 72.99 & 61.94 & 80.77 & 71.34 & 63.20 \\
         & \hspace{1.45cm} $+$ ODIN & 69.92 & 81.53 & 61.94 & 70.80 & 80.14 & 63.20 \\
         & \hspace{1.45cm} $+$ DOCTOR & 79.68 & 73.95 & 61.94 & 80.50 & 71.96 & 63.20 \\
         \cmidrule(l{3pt}r{3pt}){2-5} \cmidrule(l{3pt}r{3pt}){6-8}
         & ORCA-B & \underline{80.24} & \textbf{71.13} & 62.11 & \underline{80.77} & \textbf{69.19} & 63.02 \\
         & ORCA-R & \textbf{80.57} & \underline{72.41} & \textbf{62.29} & \textbf{80.91} & 71.70 & \textbf{63.20} \\

    \bottomrule
    \end{tabular}
    }
    \caption{Performance on \textit{CIFAR-10/100} and \textit{ImageNet}. AUROC, FPR@95TPR (FPR95), and ACC are percentages. With ACC taken into account, \textbf{bold} indicate the best results, \underline{underlined} denote ours with the second best results.}
    \label{tab:cifar}
\end{table*}

\section{Experiment} \label{sec:experiment}
\vspace{5pt}
\textbf{Datasets.}
We evaluate ORCA on a wide variety of datasets:

\noindent\textbf{1. Natural Image Benchmark}
(1) \textit{CIFAR-10/100} \cite{Krizhevsky2009LearningML} is a popular image recognition benchmark spanning across 10/100 categories.
(2) \textit{ImageNet-1K}~\cite{imagenet} a well-known benchmark in computer vision, containing 1000 fine-grained categories, with 1,281,167 training and 50,000 validation samples. This benchmark contains fine-grained categories that are visually similar, making the failure detection task more challenging.

\noindent\textbf{2. Satellite Image Benchmark}
(3) \textit{EuroSAT}~\cite{Helber2017EuroSATAN} is a satellite RGB image dataset, containing $10$ categories of land usage, such as forest, river, residential buildings, industrial buildings, \textit{etc}. The dataset comprises of 27,000 geo-referenced samples.
(4) \textit{RESISC45}~\cite{Cheng_2017} is a public benchmark for Remote Sensing Image Scene Classification. It contains 31,500 images, covering 45 scene categories with 700 images in each categories.

\vspace{5pt}
\noindent\textbf{Baselines.}
We compare ORCA to $3$ models in combination with $3$ CSFs, yielding a total of $9$ baselines. Note that we only compare with post-hoc CSFs because our methods do not require any training.

\noindent\textbf{1. Models} 
(1) \textit{Zero-shot}~\cite{Radford2021LearningTV}: The prediction of zero-shot CLIP relies on the text category name as introduced in the original paper. We compute the logits using Eq.~\ref{eq:simscore} and apply CSFs to calculate the model's confidence.
(2) \textit{Ensemble}~\cite{Radford2021LearningTV}: This model ensembles multiple templates into zero-shot classification, effectively acting as an ensemble method. We average the similarity scores from multiple templates for each category before extracting the softmax logits.
(3) \textit{DescCLIP}~\cite{Menon2022VisualCV}: As described in Sec.~\ref{sec:desc-clip}, DescCLIP averages the similarity scores of all the concepts for each category; we then apply CSFs to estimate the confidence score.

\noindent\textbf{2. CSFs}
(1) \textit{MSP}~\cite{Hendrycks2016ABF}: The confidence score is measured by taking the maximum value of the softmax responses.
(2) \textit{ODIN}~\cite{liang2020enhancing}: This CSF is a temperature-scaled version of MSP. We use the default temperature $T=1000$ and do not use perturbation for a fair comparison.
(3) \textit{DOCTOR}~\cite{Granese2021DOCTORAS}: Unlike MSP, DOCTOR fully exploits all available information contained in the soft-probabilities of the predictions to estimate the confidence.

\vspace{5pt}
\noindent\textbf{Implementation Details.} 
We utilize CLIP's ResNet-101 and ViT-B/32 backbones to perform zero-shot prediction on the benchmarks and calculate the performance metrics.
For dataset with few categories, such as \textit{CIFAR-10} and \textit{EuroSAT}, we use different prompts to retrieve diverse collections of concepts from the large language model GPT-3.5~\cite{Brown2020LanguageMA,peng2023gpt35turbo} and manually select the top $10$ visual concepts that are the most distinctive among categories. An example of our prompt is as follows, with more details in the Supplementary:
\begin{lstlisting}[breakatwhitespace=true]
Q: What are some distinctive visual concepts of [CATEGORY]?
A: Some distinctive visual concepts of [CATEGORY] are:
\end{lstlisting}
For datasets with a larger number of categories, we use the concept collection provided by Yang et al.~\cite{Yang2022LanguageIA}. 
This collection contains up to $500$ concept candidates per category; we then select the top concepts that yield the highest average similarity score with the images within each category to form $\mathcal{A}$.
We include the number of concepts used for each dataset in Table~\ref{tab:cifar} and~\ref{tab:scientific}.

\subsection{Evaluation Metrics} 
\textbf{Failure detection accuracy (AUROC).} This evaluation protocol, a threshold-independent performance evaluation, measures the area under the receiver operating characteristic curve as CSFs inherently perform binary classification between correct and incorrect predictions. A higher value denotes better ability to predict failures.

\noindent\textbf{False positive rate (FPR@95TPR).} This metric denotes the false positive rate or the probability that a misclassified sample is predicted as a correct one when the true positive rate is at 95\%.
It is a fraction that the model falsely assigns higher confidence values to incorrect samples, reflecting the tendency to be overly confident in incorrect predictions.

\noindent\textbf{Classification accuracy (ACC).} A classifier with low accuracy might produce easy-to-detect failures~\cite{jaeger2023a} and benefit from a high AUROC. Ideally, we wish a model to yield a high AUROC and ACC, and a low FPR simultaneously. 

\subsection{Results on Natural Image Benchmarks}\label{subsec:natural}
We report the performance of all methods on the three evaluation metrics on the natural image benchmarks on ResNet-101 and ViT-B/32 and provide the following \textit{observations}:
\begin{tcolorbox}[width=\linewidth,colback={verylightgray}, colframe={lightgray},left=0pt, right=0pt, top=0pt, bottom=0pt]
{\bf Observation 1:} {Concept-based methods demonstrate better failure detection.}
\end{tcolorbox}
\noindent Table~\ref{tab:cifar} shows DescCLIP and ORCA consistently achieves higher AUROC compared to Zero-shot and Ensemble, especially on datasets with a large number of categories, such as \textit{CIFAR-100} and \textit{ImageNet}. 
The augmentation to multiple signals per category helps concept-based methods obtain a finer-grained analysis for better failure detection.
On a different note, Ensemble boosts the Zero-shot's ACC but still results in a lower AUROC and higher FPR on the large-scale datasets for both backbones.
Ensemble, in the same principles as concept-based methods, augments the number of signals; however, we hypothesize the \textit{lack of diversity} in those signals deteriorates the separability between correct and incorrect samples.
\begin{tcolorbox}[width=\linewidth,colback={verylightgray}, colframe={lightgray},left=0pt, right=0pt, top=0pt, bottom=0pt]
{\bf Observation 2:} {Our method reduces overconfident but incorrect predictions.}
\end{tcolorbox}
\noindent In Table~\ref{tab:cifar}, we observe that our methods consistently reduce the false positive rate across datasets and for both backbones.
Both variants of ORCA decrease the FPR@95TPR substantially while keeping AUROC and ACC competitive.
On \textit{ImageNet}, ORCA-B achieves the best performance on this metric, outperforming the zero-shot model and DescCLIP by $3.72\%$ and $2.15\%$ respectively using ViT-B/32.
We hypothesize that allowing the model to recognize an object from different angles provides more reliable confidence assessment, enabling faithful failure detection while also achieving superior predictive accuracy.

\subsection{Results on Satellite Image Benchmarks}
We report the performance on \textit{EuroSAT} and \textit{RESISC45} on ResNet-101 and ViT-B/32. Note that all results are zero-shot performance. We discuss the following \textit{observation}:
\begin{tcolorbox}[width=\linewidth,colback={verylightgray}, colframe={lightgray},left=0pt, right=0pt, top=0pt, bottom=0pt]
{\bf Observation 3:} {Our method boosts both predictive and failure detection accuracy on remote sensing benchmarks.}
\end{tcolorbox}
\noindent Table~\ref{tab:scientific} shows that ORCA-R consistently outperforms all baselines on all evaluation metrics.
Compared to DescCLIP $+$ MSP on \textit{EuroSAT}, ORCA-R enjoys a $3.6\%$ improvement in AUROC and $6.25\%$ in FPR while boosting the overall accuracy by $1.49\%$. On \textit{RESISC45}, while ORCA-R's improvement on AUROC and ACC is marginal, it significantly reduces FPR. Additionally, these datasets represent out-of-distribution data for CLIP, underscoring ORCA's enhanced reliability and robustness against such distributional variations.

\begin{table*}[!htb]
    \centering
    \resizebox{0.8\linewidth}{!}{
    \begin{tabular}{llcccccc}
    \toprule
          \multirow{2}{*}{\textbf{Dataset}}&\multirow{2}{*}{\textbf{Method}}& \multicolumn{3}{c}{ResNet-101} & \multicolumn{3}{c}{ViT-B/32}\\
          \cmidrule(l{3pt}r{3pt}){3-5} \cmidrule(l{3pt}r{3pt}){6-8}
          &&\textbf{AUROC} $\uparrow$ & \textbf{FPR95} $\downarrow$ & \textbf{ACC} $\uparrow$ & \textbf{AUROC} $\uparrow$ & \textbf{FPR95} $\downarrow$ &\textbf{ACC} $\uparrow$  \\
          
    \cmidrule(l{3pt}r{3pt}){1-5} \cmidrule(l{3pt}r{3pt}){6-8}
         \multirow{7}{2cm}{EuroSAT ($K = 10$)}& Zero-shot $+$ MSP & 61.73& 88.98 & 30.30 & 76.42& 80.24& 41.11\\
         & \hspace{1.35cm} $+$ ODIN & 61.35 & 89.38 & 30.30 & 75.54 & 79.28 & 41.11\\
         & \hspace{1.35cm} $+$ DOCTOR & 60.76 & 89.85 & 30.30 & 76.67 & 79.30 & 41.11\\
         & Ensemble $+$ MSP & 54.69 & 92.21 & 31.90 & 66.83 & 89.19 & 48.73\\
         & \hspace{1.4cm} $+$ ODIN & 55.10 & 93.09 & 31.90 & 65.73 & 90.09 & 48.73\\
         & \hspace{1.4cm} $+$ DOCTOR & 53.73 & 94.09 & 31.90 & 61.14 & 90.63 & 48.73\\
         & DescCLIP $+$ MSP & 64.89 & 86.39 & 33.13 & 73.93&  77.54 & 48.51\\
         & \hspace{1.45cm} $+$ ODIN & 64.16 & 87.16 & 33.13 & 71.74 & 78.34 & 48.51 \\
         & \hspace{1.45cm} $+$ DOCTOR & 62.79 & 89.05 & 33.13 & 72.74 & 79.85 & 48.51 \\
         \cmidrule(l{3pt}r{3pt}){2-5} \cmidrule(l{3pt}r{3pt}){6-8}
         & ORCA-B & \underline{67.86} & \textbf{86.43} & \underline{34.11} &  76.20&  77.80 & \underline{49.74}\\
         & ORCA-R & \textbf{69.01} & \textbf{86.43} & \textbf{34.76} &  \textbf{77.55}&  \textbf{71.29}& \textbf{50.00}\\

    \cmidrule(l{3pt}r{3pt}){1-5} \cmidrule(l{3pt}r{3pt}){6-8}
         \multirow{7}{2cm}{RESISC45 ($K = 10$)}& Zero-shot $+$ MSP & 68.13 & 87.04 & 37.66 & 77.92 & 80.35 & 55.57 \\
         & \hspace{1.35cm} $+$ ODIN & 62.60 & 89.48 & 37.66 & 71.66 & 84.85 & 55.57 \\
         & \hspace{1.35cm} $+$ DOCTOR & 67.57 & 87.19 & 37.66 & 76.95 & 82.17 & 55.57 \\
         & Ensemble $+$ MSP & 68.87 & 85.39 & 39.79 & 78.40 & 80.14 & 56.68 \\
         & \hspace{1.4cm} $+$ ODIN & 62.67 & 89.57 & 39.79 & 71.99 & 85.31 & 56.68 \\
         & \hspace{1.4cm} $+$ DOCTOR & 67.88 & 87.29 & 39.79 & 77.54 & 82.34 & 56.68 \\
         & DescCLIP $+$ MSP & 73.44 & 79.78 & 43.16 & 77.47 & 82.25 & 58.33 \\
         & \hspace{1.45cm} $+$ ODIN & 69.47 & 84.61 & 43.16 & 71.49 & 86.21 & 58.33 \\
         & \hspace{1.45cm} $+$ DOCTOR & 72.95 & 79.89 & 43.16 & 76.81 & 84.88 & 58.33 \\
         \cmidrule(l{3pt}r{3pt}){2-5} \cmidrule(l{3pt}r{3pt}){6-8}
         & ORCA-B & 71.88 & 90.41 & \textbf{46.22} & \underline{77.71} & 86.31 & 59.10 \\
         & ORCA-R & \textbf{74.28} & \textbf{80.31} & \underline{45.13} & \textbf{78.24} & \textbf{76.52} & \textbf{59.10}\\
    
    \bottomrule
    \end{tabular}
    }
    \caption{Performance on \textit{EuroSAT} and \textit{RESICS45}. AUROC, FPR@95TPR (FPR95), and ACC are percentages. With ACC taken into account, \textbf{bold} indicate the best results, \underline{underlined} denote ours with the second best results.}
    \label{tab:scientific}
\end{table*}

\subsection{Ablation Studies}~\label{ablate}
We conduct two ablation studies on the effect of the number of concepts and the choice of the weighting function used for ORCA-R in this section.
\begin{figure}[!ht]
  \centering
  
     \includegraphics[width=.75\linewidth]{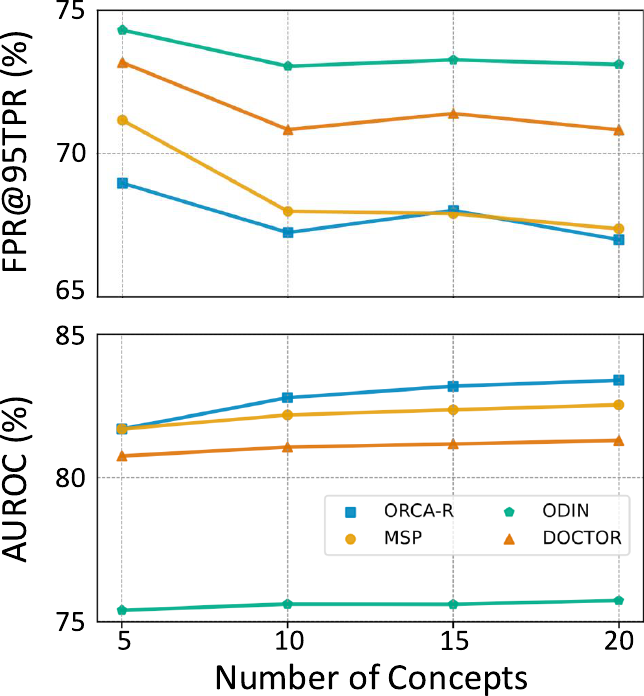}
     \caption{Failure detection accuracy (AUROC) and false positive rate (FPR@95TPR) across different numbers of concepts on \textit{CIFAR-100}. Overall, we can an increase in the number of concepts boosts the performance in both metrics.}
    \label{fig:no-concepts}
\end{figure}

\begin{figure}[!ht]
  \centering
         \includegraphics[width=0.65\linewidth]{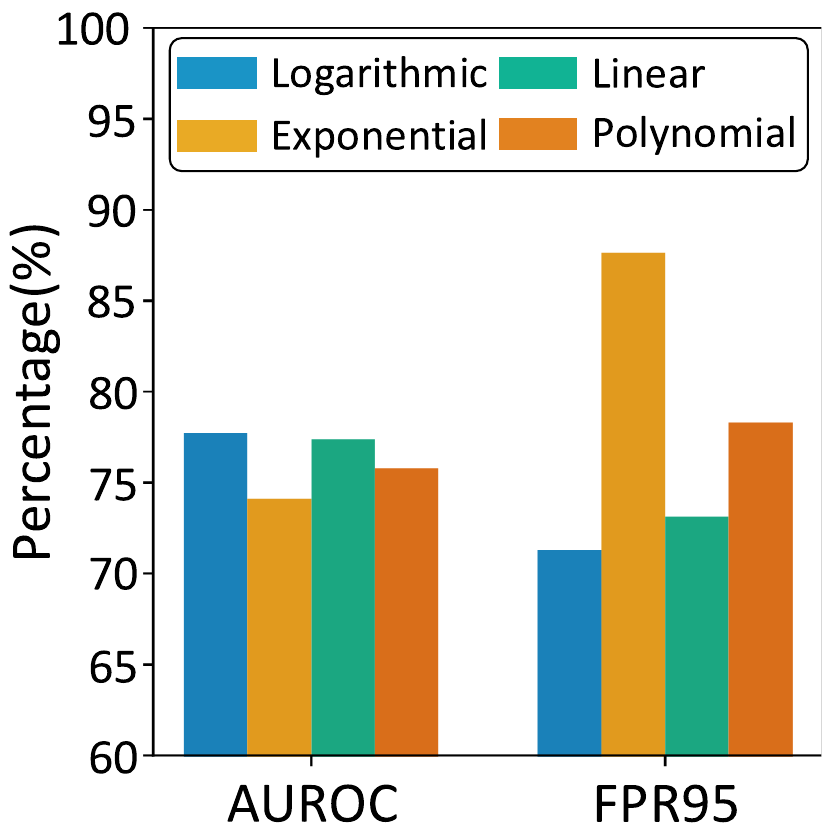}
         \caption{
         Failure detection capabilities of each weighting function on \textit{EuroSAT}, where \texttt{Logarithmic} consistently outperforms others.}
         \label{fig:wf}
\end{figure}


\noindent\textbf{Ablation on number of concepts.}
We use the ViT-B/32 backbone on \textit{CIFAR-100} and $K=\{5, 10, 15, 20\}$ for this experiment. We study the effect of the number of concepts on the performance on AUROC and FPR@95TPR of DescCLIP $+$ MSP, ODIN, DOCTOR and ORCA-R.
Fig.~\ref{fig:no-concepts} shows that the FPR of ORCA-R is consistently lower than those of the other baselines across various $K$.
We also see an increasing (decreasing) trend in AUROC (FPR) as the number of concepts rises. This signifies a finer-grained assessment both enables better failure detection and alleviates the problem of assigning high confidence to incorrect predictions.

\noindent\textbf{Ablation on choice of weighting function.}
We examine how various weighting functions influence the failure detection efficacy of ORCA-R. Fig.~\ref{fig:wf} (left) visualizes the weight distribution on the top-$10$ concepts among the weighting functions. 
In Figure~\ref{fig:wf} (right), \texttt{Logarithmic} outperforms others, contrasting with \texttt{Exponential}, which exhibits the least effectiveness.
\texttt{Logarithmic} ensures a balanced distribution of weights, recognizing the importance of higher-ranked concepts while also accounting for lower-ranked ones.
Conversely, \texttt{Exponential} significantly overweighs the highest-ranked concept, neglecting the contributions of those ranked lower.
\section{Failure Interpretation} \label{sec:interpretation}

\begin{figure}
    \centering
    \begin{subfigure}[b]{\linewidth}
         \centering
         \includegraphics[width=\linewidth]{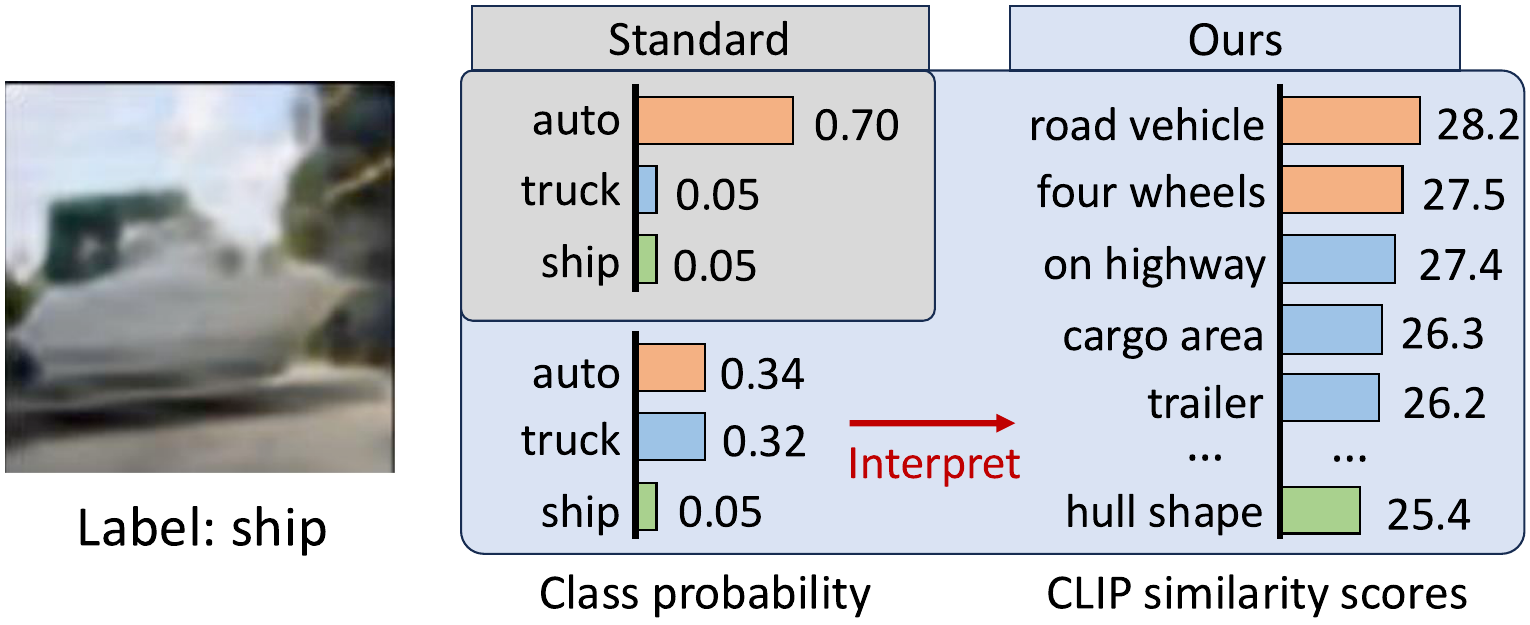}
         \caption{Failure caused by spurious correlation.}
         \label{viz:ship-land}
         \vspace{0.2cm}
     \end{subfigure}
    
     \begin{subfigure}[b]{\linewidth}
         \centering
         \includegraphics[width=\linewidth]{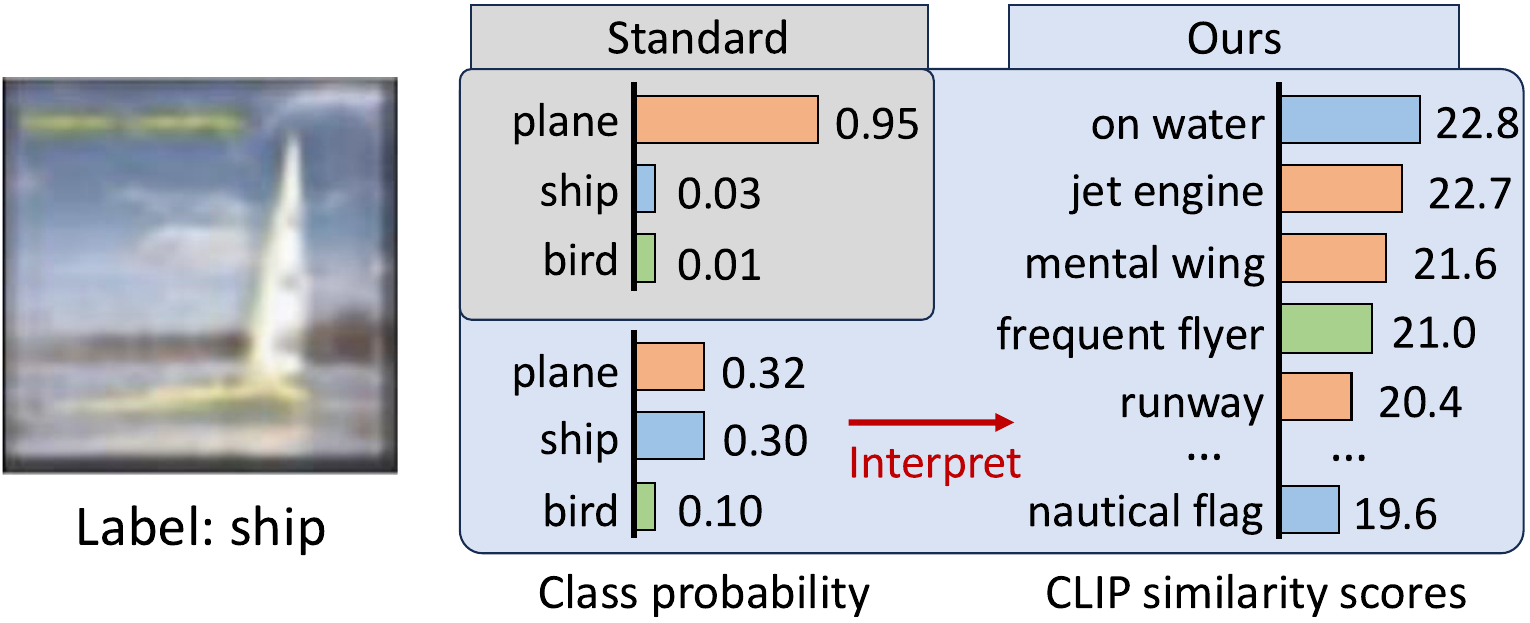}
         \caption{Failure caused by cross-category resemblance.}
         \label{viz:ship-sky}
     \end{subfigure}
    \caption{Failure interpretation with human-level concepts. We show the confidence scores of the top $3$ categories (left histograms) and similarity scores of the top $10$ concepts (right histograms) from \textit{CIFAR-10}. Standard methods might output overconfident misclassifications due to: (a) \textit{spurious correlation} and (b) \textit{cross-category resemblance}. Concept-level signals not only achieves better failure detection capability in such scenarios but also enables further interpretation of \textit{why} the model fails. ``auto" is short for ``automobile."}
    \label{fig:visual}
\end{figure}

ORCA not only achieves superior failure detection but also enables failure interpretation with human-level concepts.
We discuss two scenarios that cause the model to output overconfident values on misclassified samples: \textit{spurious correlation} and \textit{cross-category resemblance} (Fig.~\ref{fig:visual}). 

In the former scenario (Fig.~\ref{viz:ship-land}), the presence of a road (a spurious feature) leads the model to misclassify the ship as a land vehicle, automobile or truck. 
We demonstrate that a standard model struggles to identify such failures, resulting in a high confidence score for automobile. 
In contrast, ORCA leverages human-level concepts, offering more nuanced signals for a refined assessment of the model's confidence. 
For instance, strong responses from concepts like ``road vehicle" and ``four wheels" for automobile, and ``cargo area" and ``trailer" for truck, contribute to a significantly lower confidence. 
Furthermore, we can easily interpret \textit{why} the model makes such a prediction through concepts.

In the latter scenario (Fig.~\ref{viz:ship-sky}), the ship (sailboat) bears a resemblance to an airplane from a distance. The similarity between the sky and water also creates an illusion of the object being airborne. The top-$K$ concepts from our method exhibit strong responses to concepts associated with airplanes and birds. Analyzing this information allows us to confidently deduce that the model misclassifies the image as an airplane due to the sky-like background and the object's resemblance to an airplane.

\begin{links}
\link{Code}{https://github.com/Nyquixt/ORCA}
\end{links}

%

\section*{Acknowledgments}
This work is supported by the DoD DEPSCoR Award AFOSR FA9550-23-1-0494, the  NSF CAREER Award No. 2340074, the NSF SLES Award No. 2416937, and the NSF III CORE Award No. 2412675.
Any opinions, findings and conclusions or recommendations expressed in this material are
those of the authors and do not reflect the views of the supporting entities.

\bibliography{aaai25}

\end{document}